\documentclass[conference]{IEEEtran}
\IEEEoverridecommandlockouts
\usepackage{amsmath,amssymb,amsfonts}
\usepackage{algorithmic}
\usepackage{graphicx}
\usepackage{textcomp}
\usepackage{xcolor}
\usepackage{amsmath, xparse}
\usepackage{float}
\usepackage{biblatex}
\usepackage{multirow}
\usepackage{graphicx}
\usepackage{makecell}
\usepackage{booktabs}
\usepackage{array}
\usepackage{hyperref}
\bibliography{citations}
\def\BibTeX{{\rm B\kern-.05em{\sc i\kern-.025em b}\kern-.08em
    T\kern-.1667em\lower.7ex\hbox{E}\kern-.125emX}}
\begin{document}
\title{Physics-Guided Reinforcement Learning System for Realistic Vehicle Active Suspension Control\\
}

\makeatletter
\newcommand{\linebreakand}{%
  \end{@IEEEauthorhalign}
  \hfill\mbox{}\par
  \mbox{}\hfill\begin{@IEEEauthorhalign}
}
\makeatother

\author{
    \IEEEauthorblockN{Anh N. Nhu}
    \IEEEauthorblockA{
        \textit{Department of Computer Science} \\
        \textit{University of Maryland, College Park}\\
        MD 20742, USA\\
        anhu@terpmail.umd.edu
    }
\and
    \IEEEauthorblockN{Ngoc-Anh Le}
    \IEEEauthorblockA{
        \textit{Institute of Mechanical Engineering} \\
        \textit{University of Transport and Communications}\\
        Hanoi, Vietnam\\
        anh181300098@lms.utc.edu.vn
    }
\and
\IEEEauthorblockN{Shihang Li}
    \IEEEauthorblockA{
        \textit{Department of Computer Science} \\
        \textit{University of Michigan, Ann Arbor}\\
        MI 48109, USA\\
        shihangl@umich.edu
    } 
\linebreakand
\IEEEauthorblockN{Thang D.V. Truong}
    \IEEEauthorblockA{
        \textit{School of Mechanical Engineering} \\
        \textit{Hanoi University of Science and Technology}\\
        Hanoi, Vietnam\\
        thang.truongdangviet@hust.edu.vn
    }
}

\maketitle

\begin{abstract}
The suspension system is a crucial part of the automotive chassis, improving vehicle ride comfort and isolating passengers from rough road excitation. Unlike passive suspension, which has constant spring and damping coefficients, active suspension incorporates electronic actuators into the system to dynamically control stiffness and damping variables. However, effectively controlling the suspension system poses a challenging task that necessitates real-time adaptability to various road conditions. This paper presents the Physics-Guided Deep Reinforcement Learning (DRL) for adjusting an active suspension system's variable kinematics and compliance properties for a quarter-car model in real time. Specifically, the outputs of the model are defined as actuator stiffness and damping control, which are bound within physically realistic ranges to maintain the system's physical compliance. The proposed model was trained on stochastic road profiles according to ISO 8608 standards to optimize the actuator's control policy. According to qualitative results on simulations, the vehicle body reacts smoothly to various novel real-world road conditions, having a much lower degree of oscillation. These observations mean a higher level of passenger comfort and better vehicle stability. Quantitatively, DRL outperforms passive systems in reducing the average vehicle body velocity and acceleration by 43.58\% and 17.22\%, respectively, minimizing the vertical movement impacts on the passengers. The code is publicly available at \href{https://github.com/anh-nn01/RL4Suspension-ICMLA23}{github.com/anh-nn01/RL4Suspension-ICMLA23}.
\end{abstract}


\section{Introduction}

As a vital component of vehicles,  the suspension system plays a significant role in improving passenger comfort, driving stability, and protection of other components. Mechanically, the suspension system consists of essential components including springs, bump stops, shock absorbers, and arms or links connecting the vehicle chassis frame \cite{THEUNISSEN2021206}. During vehicle motion, when encountering bumps caused by stochastic road excitation, the suspension system serves a crucial function in effectively isolating vibrations, minimizing the impact experienced by passengers, and enhancing overall vehicle stability.

In the area of controlled systems, there are three major categories of suspension systems: passive, semi-active, and active suspension systems. The passive suspension system has a fixed spring and shock absorber. However, since the road environment constantly changes and is unpredictable, using a passive suspension system is unable to maintain the best vehicle operation experience. Given that the passive suspension system has fixed dampers and does not allow external control, it fails to handle possible road disturbance or fully absorb vibrations.\cite{review_suspension} Therefore, active and semi-active suspension systems with adjustable operational parameters were introduced to overcome the weaknesses of passive suspension systems. They allow built-in electronic control units to manipulate multiple parameters such as damping rates\cite{doi:10.1177/1461348419876392} and additional forces on spring\cite{10.6703--570436548} based on different road situations in real-time. 

The previous controllers including proportional–integral–derivative (PID), linear parameter varying (LPV), multi-objective control strategy MCTCS~\cite{10072041}, and Linear–quadratic–Gaussian (LQG) \cite{10.1155/2023/2889435} have been the common and conventional control methods. However, the most prominent limitation of these approaches is that their performance is highly sensitive to parameter configuration, which relies heavily on domain expertise and vehicle specifications (e.g.: vehicle weight). Therefore, the parameters are not generalizable to new vehicles and demand extensive time and effort to find the optimal set of parameters. Furthermore, the above algorithms are all linear controllers while active suspension system in real practice has non-linear and complex characteristics \cite{MOHITE20184317}. To be more specific, the road condition is stochastic, resulting in highly non-linear vehicle vibrations and making the linear control effect less effective.

In the context of computer science and vehicle controls, researchers have been proposing different models and control strategies to improve the live adjustment of active suspension system parameters for a better driving experience. Deep reinforcement learning (DRL) focuses on learning optimal actions in a sequential decision-making process. Controllers based on the reinforcement learning model enable learning from the interaction with the environment to adapt the suspension system parameters under certain road conditions leading to improved vehicle overall performance. Compared to Supervised Deep Learning methods \cite{5459681}, DRL does not require human-labeled or controller-generated samples for training and can generate large amounts of training data solely by physics simulation, which inspires many existing works on DRL-based controllers. \cite{8606740} utilized the actor-critic scheme and compared the model performance to a conventional quarter suspension system to demonstrate its feasibility. On the other hand, Xi Wang \emph{et~al.} adopted the deep Q Network (DQN) \cite{9442091} technique to the quarter suspension system and achieved higher driving comfort compared to Skyhook Damper Control (SDG). Nonetheless, many previous works are limited to experimenting with idealized and simplified road profiles, which do not reflect continuous variations of real-world roads. Most simulation results of these studies show a promising new generation of control strategies. However, some results show that the system response is quite fast according to random rough roads, which causes discomfort to passengers. In contrast, others have smoother system responses, yet the delay is too significant, causing inconsistency in dynamic control. Another major DRL technique, Deep Deterministic Policy Gradient (DDPG) was introduced by \cite{8950280} to control the semi-active quarter suspension system. Specifically, \cite{8950280} proposed to employ DDPG to learn optimal control strategy for a vertical force $u_F \in [-600 N, 600 N]$ exerted on the system. Their approach yields promising control results on ISO road class D and E, demonstrating the potential usage of DRL for suspension control problems. However, from the mechanical point of view, this force is vaguely defined, and the embedded hardware to generate such force is not clearly presented. In other words, how to physically exert the computed force is out of the scope of their study. Therefore, this force can be hard to be applied in real physical systems, reducing the viability of this approach. 

To address the aforementioned limitations of existing works, a Physics-Guided Reinforcement Learning framework is proposed to obtain an optimal, physically practical active suspension control system. Specifically, the outputs of the DRL Policy Network are explicitly constrained to two specific variables: active stiffness $(k_a)$ and active damping ($c_a$). Controlling $k_a$ and $c_a$ is physically feasible, which is explained in Section~\ref{sec:QuarterModel}. Although the force is not directly controlled, in practice, dynamically controlled $k_a$ and $c_a$ results in a force on the system described in Equation~\ref{eq:f_a}. In summary, compared to prior work, our proposed active control approach is more realistic and physically well-defined. The DRL controller is tested on several ISO road profiles for evaluation, outperforming the passive control system. 



\section{Vehicle Suspension and Road Excitation Model}\label{sec:QuarterModel}
In this paper, a two-degree-of-freedom (DOF) quarter-car model has been implemented to evaluate the capabilities of active suspension systems. Quarter-car model is designed to represent the vertical dynamic of a car. The goal of active suspension is to minimize vertical acceleration. The quarter car model is suitable for evaluating the performance of various control strategies\cite{handbook}. It consists of four parts: a sprung mass, a suspension component, an unsprung mass, and a wheel. Active suspension utilizes separate actuators to exert independent forces on the existing suspension to improve the vehicle's riding characteristics, comfort, and handling by actively adjusting the suspension in response to road conditions or driving dynamics. In practical implementation, this actuator is controlled with the use of hydraulics or pneumatics.

\begin{figure}[htbp]
\centerline{\includegraphics[scale=0.4]{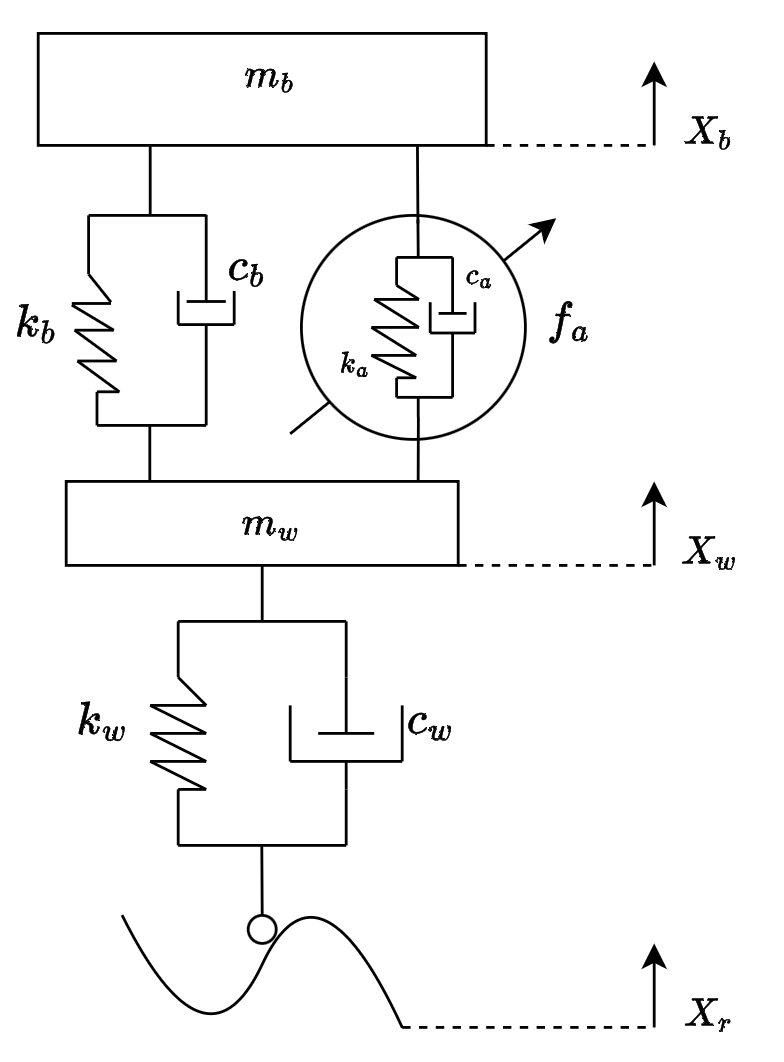}}
\caption{Quarter vehicle model with active suspension}
\label{fig:Quarter-Model}
\end{figure}

Figure~\ref{fig:Quarter-Model} illustrates the quarter car model with active suspension system where $m_b$ is the quarter body mass (or sprung mass) (kg), $m_w$ is the wheel mass (or unsprung mass) (kg), $k_b$ is the suspension spring stiffness (N/m), $c_b$ is the suspension damping coefficient (Ns/m), $k_w$ is wheel stiffness (N/m), $c_w$ is the wheel damping coefficient (Ns/m), $f_a$ is the active control force generated by the linear motor actuator, $k_a$ is the active (controllable) stiffness (N/m), $c_a$ is the active (controllable) damping coefficient (Ns/m), $x_b$ and $x_w$ are the absolute displacement of the body and wheel respectively (m), $x_r$ is the road excitation (m).


The dynamic equations of the system which satisfy Newton’s second law can be provided by the following equations:\\
\begin{equation}
    m_b \ddot{x}_b = -k_b (x_b - x_w) - c_b (\dot{x}_b - \dot{x}_w) + f_a
\end{equation}
\begin{equation}
    \begin{split}
        m_w \ddot{x}_w = k_b (x_b - x_w) + c_b (\dot{x}_b - \dot{x}_w) - \\k_w (x_w - x_r) - c_w (\dot{x}_w - \dot{x}_r) - f_a
    \end{split}
\end{equation}

The state-space representation of the suspension system can be expressed as: 
\begin{equation} 
    \begin{cases}
        \mathbf{\dot{x}} = \mathbf{A}\mathbf{x} +\mathbf{B} \mathbf{U}\\
        \mathbf{y} = \mathbf{C}\mathbf{x} + \mathbf{D}\mathbf{U}
    \end{cases}
\end{equation}
\begin{equation*}
\mathbf{A} = \begin{bmatrix}
0 & 1 & 0 & 0\\ 
\frac{-k_b}{m_b} & \frac{-c_b}{m_b} & \frac{k_b}{m_b} & \frac{c_b}{m_b}\\
0 & 0 & 0 & 1\\
\frac{k_b}{m_w} & \frac{c_b}{m_w} & \frac{-(k_b + k_w)}{m_w} & \frac{-(c_b + c_w)}{m_w}\\
\end{bmatrix} \quad\\
\mathbf{B} = \begin{bmatrix}
    0 & 0\\ 
    0 & \frac{1}{m_b}\\
    0 & 0\\ 
    \frac{-k_w}{m_w} & \frac{-1}{m_w} 
\end{bmatrix}
\end{equation*}

\hfill

$\mathbf{C}= \begin{bmatrix}
   1 & 0 & -1 & 0\\
   \frac{-k_b}{m_b} & \frac{-c_b}{m_b} & \frac{k_b}{m_b} & \frac{c_b}{m_b}
\end{bmatrix}$ \quad
$\mathbf{D} = \begin{bmatrix}
   0 & 0\\
   0 & \frac{1}{m_b}
\end{bmatrix}$

\hfill

\hfill

$\mathbf{x} = \begin{bmatrix}
    x_b\\
    \dot{x_b}\\
    x_w\\
    \dot{x_w}
\end{bmatrix}$ \quad
$\mathbf{y} = \begin{bmatrix}
    x_b - x_w\\
    \ddot{x_b}
\end{bmatrix}$ \quad
$\mathbf{U} = \begin{bmatrix}
    x_r\\
    f_a
\end{bmatrix}$ \quad \\

The tire damping can be neglected due to its relatively insignificant magnitude compared to the damping coefficient of shock absorber.



\begin{figure*}[htbp]
  \centering
  \begin{tabular}{cc}
    \includegraphics[width=\linewidth]{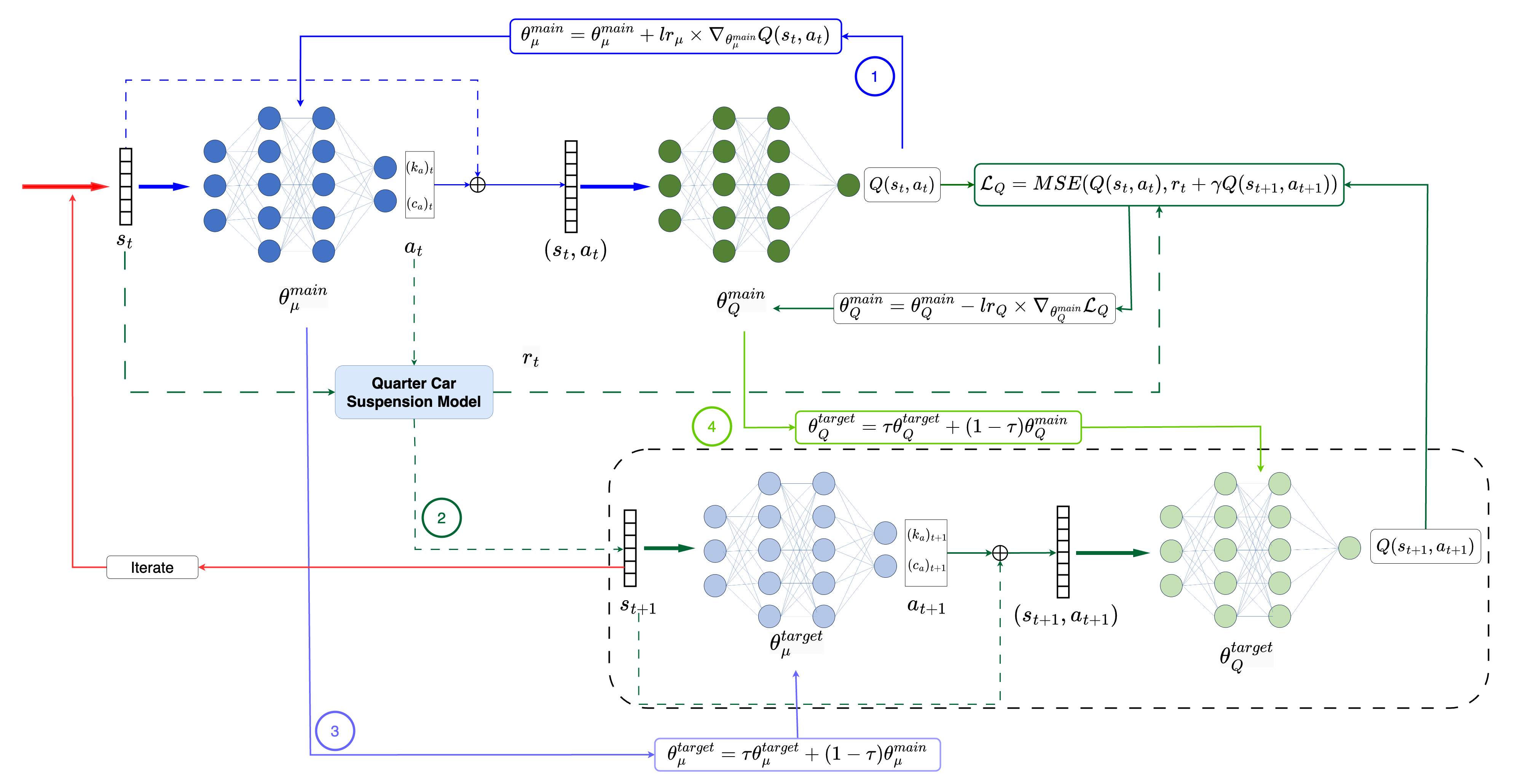} &
  \end{tabular}
  \caption{Algorithmic Flow of the DDPG agent Training Pipeline for Active Suspension Control at each time step. Numbers 1 to 4 denote the chronological update order of the networks in each time step. Specifically, the networks shorted by updating order are: (1) Main Policy Network (SGD), (2) Main Q-Network (SGD), (3) Target Policy Network (soft update), and (4) Target Q-Network (soft update).}
  \label{fig:AlgoFlow}
\end{figure*}

\section{Deep Deterministic Policy Gradient (DDPG) for Active Suspension Control}

Using the 2-DOF quarter suspension system specified in \cite{ISO8608}, a physical environment is simulated to train the Active Suspension DRL agent, specifically to model state transition and compute the reward function. There are 6 fixed parameters in this 2-DOF model: $m_b$, $m_w$, $k_w$, $c_w$, $k_b$, $c_b$ representing the physical properties of the vehicle's masses, the tire's stiffness, and the fixed parallel suspension system. The outputs of Policy Network are active stiffness $k_a$ and active damping $c_a$, which are bounded for realistic control in practice. Specific ranges of all mechanical parameters are reported in Table~\ref{table:hyperparams}. In practice, the dynamical control of $k_a$ and $c_a$ results in a force $f_a$, displayed in Figure~\ref{fig:Quarter-Model}. The formula of $f_a$ as a function of DRL-controlled $k_a$ and $c_a$ is:
\begin{equation}\label{eq:f_a}
    f_a = (k_b + k_a) \cdot x_b + ({c_b} + {c_a}) \cdot \dot{x_b}
\end{equation}

In this paper, the Deep Deterministic Policy Gradient (DDPG) algorithm is employed \cite{1509.02971} to obtain optimal dynamic control policy for the active suspension system. The general algorithmic flow of the framework is shown in Figure~\ref{fig:AlgoFlow}. The DDPG framework has an Actor-Critic architecture that consists of two neural networks: Policy Network (Actor) and Q-Network (Critic). The Policy Network learns the suspension control policy, mapping the observed state to optimal actions ($k_a$ and $c_a$). The Q-Network is responsible for estimating the Q-value (expected cumulative reward) of the state-action pairs. To increase the stability of the training process, the soft update technique is used \cite{KOBAYASHI202163}. This soft update helps to reduce abrupt changes in the target value estimation during the training process. Instead of updating the Target Networks at each time step, soft update slowly tracks the Main Networks (updated at each time step) and updates the Target Networks using the weighted average of the Main Networks' weights:

\begin{equation}
    \theta^{target}_{Q} = \tau \theta^{target}_{Q} + (1-\tau)\theta^{main}_{Q}
\end{equation}
\begin{equation}
    \theta^{target}_{\mu} = \tau \theta^{target}_{\mu} + (1-\tau)\theta^{main}_{\mu}
\end{equation}
$\theta^{target}_{Q}$ and $\theta^{main}_{Q}$ are the weights of Target and Main Q-Network, $\theta^{target}_{\mu}$ and $\theta^{main}_{\mu}$ are the weights of Target and Main Policy Network, and $\tau \in [0,1]$ is a hyperparameter that controls the rate at which the target network's weights track the main network's weights. Similar to DQN \cite{DBLP:journals/corr/MnihKSGAWR13}, DDPG utilizes experience replay to store past experiences (state, action, reward, next state) in a replay buffer. At each time step, a batch of experiences is randomly sampled from the replay buffer to decorrelate the data, increase the model's exposure to uncommon scenarios, and make the learning more stable. During the training process, exploration noise ($\sigma \in [0,1]$) is added to the Policy Network to facilitate discoveries of more optimal control policies. The exploration noise is gradually reduced from $0.5$ to $0.05$ as the number of training episodes increases. However, this noise is disabled during the inference process to output deterministic actions. The optimal set of hyper-parameters, including model architecture, $\tau$, $\sigma$, $\gamma$ are reported in Table~\ref{table:hyperparams}.
Multiple existing works on DRL-based Semi-active suspension system used $s_{\mu} = \begin{bmatrix} x_b & x_w & \dot{x_b} & \dot{x_w} \end{bmatrix}^T$ as the input state space (\cite{8950280, shen2022, babaahmadi2023deep}). In our work, $x_b$ and $x_w$ are removed as this removal helps speed up the optimization convergence while empirically increasing the performance of the agent on novel road profiles. The input state space of the Policy Network at each time step $t$ is:
$$(s_{\mu})_t = 
\begin{bmatrix}
    \dot{x_b}_{(t)} \\
    \dot{x_w}_{(t)} \\
    \dot{x_r}_{(t)} \\
    \dot{x_b}_{(t-1)} \\
    \dot{x_w}_{(t-1)} \\
    \dot{x_r}_{(t-1)}
\end{bmatrix} \in \mathbb{R}^6$$

Additionally, information is introduced from past time step $\dot{x_b}_{(t-1)}, \dot{x_w}_{(t-1)}, \dot{x_r}_{(t-1)}$ to the state space to exploit potential temporal dependency. The output of the Policy Network at each time step $t$ is:  
$$a_t = \begin{bmatrix} ({k_a})_t  & ({c_a})_t\end{bmatrix}^T \in \mathbb{R}^2$$
where $({k_a})_t \in [-2500, 5000]$ and $({c_a})_t \in [-600, 600]$. To impose these bounds on the Policy Network's outputs while still keeping the targets differentiable, the tanh activation is used at the end of the Policy Network and scale the output linearly according to the targets' ranges. The input state space to the Q-Network is the concatenation of $(s_{\mu})_t$ and $a_t$. Q-Network outputs a single scalar value representing the expected cumulative reward when the action $a_t$ is taken from the current state $(s_{\mu})_t$. The state transition from $(s_{\mu})_t$ to $(s_{\mu})_{t+1}$ is computed using the Quarter Car Model described in Section~\ref{sec:QuarterModel}, creating a continuous feedback environment for training the agent.

The primary purpose of the active suspension system is to minimize vehicle movement impacts on passengers. We need to minimize vertical body displacement $\dot{x_b}$ and vertical body acceleration $\ddot{x_b}$ to achieve this purpose. Empirically, we find that using only $\dot{x_b}$ in the reward function ($r = f(\dot{x_b})$) yields better results in terms of both $|\dot{x_b}|$ and $|\ddot{x_b}|$ minimizations than any other combinations with $\ddot{x_b}$ ($r = f(\dot{x_b}, \ddot{x_b})$) or other variables. Our immediate reward function is $r_t$ = $-0.1|\dot{x_b}_{(t)}|$. 

The agent's weights was trained using a single NVIDIA Tesla A100-40GB GPU. The training process took approximately 2.5 hours for the Policy Network to converge. Figure~\ref{fig:Rewards} shows the cumulative reward curve of the DDPG agent of each training episode. The trained Policy Network was evaluated on novel road profiles unseen during the training process to test its performance in unfamiliar real-world situations. The results and analysis are included in Section~\ref{sec:Results}.


\begin{table}[htbp]
    \centering
    \caption{Quarter Car Parameters \& DDPG Hyper-parameters}
    \begin{tabular}{|c|c|}
        \hline
         \textbf{Quarter Car Parameters} & Value \\
         \hline
         Quarter body mass ($m_b$) & $450$ [kg] \\
         \hline
         Wheel mass ($m_w$) & $45$ [kg]\\
         \hline
         Active (controllable) stiffness ($k_a$) & $-2500$ to $5000$ [N/m]\\
         \hline
         Active (controllable) damping ($c_a$) & $-600$ to $600$ [Ns/m]\\
         \hline
         Suspension stiffness ($k_b$) & $15000$ [N/m]\\
         \hline
         Suspension damping ($c_b$) & $1500$ [Ns/m]\\
         \hline
         Wheel stiffness coefficient ($k_w$) & $150000$ [N/m]\\
         \hline
         Wheel damping coefficient ($c_w$) & $0$ [Ns/m]\\
         \hline
         \hline
         \textbf{Network Configuration} & Value \\
         \hline
         Input dimension (Q-Network) & 8-D (observation + action) \\
         Input dimension (Policy Network) & 6-D (observation) \\
         Hidden Layers (Q-Network) & $32\times32$ (2 layers) \\
         Hidden Layers (Policy Network) & $16\times16$ (2 layers) \\
         \hline
         \hline
         \textbf{Reward function} & $r_t$ = $-0.1|\dot{x_b}_{(t)}|$ \\
         \hline
         \hline
         \textbf{Training Hyper-parameters} & Value \\
         \hline
         Number of Episodes & 700 \\
         Batch size & 512 \\
         Buffer Size & $10^5$ \\
         Optimizer (Q-Network) & Adam \\
         Optimizer (Policy Network) & Adam \\
         Learning rate (Q-Network) & $10^{-3}$ \\
         Learning rate (Policy Network) & $10^{-4}$ \\
         Discount factor ($\gamma$) & 0.95 \\
         Soft Update rate ($\tau$) & 0.99 \\
         Exploration noise: Episode 1 - 100 ($\sigma_1$) & 0.5 \\
         Exploration noise: Episode 100 - 200 ($\sigma_2$) & 0.3 \\
         Exploration noise: Episode 200 - 500 ($\sigma_3$) & 0.15 \\
         Exploration noise: Episode 500 - 700 ($\sigma_4$) & 0.05 \\
     \hline
    \end{tabular}
    \label{table:hyperparams}
\end{table}

\begin{figure}[htbp]
    \centerline{\includegraphics[scale=0.4]{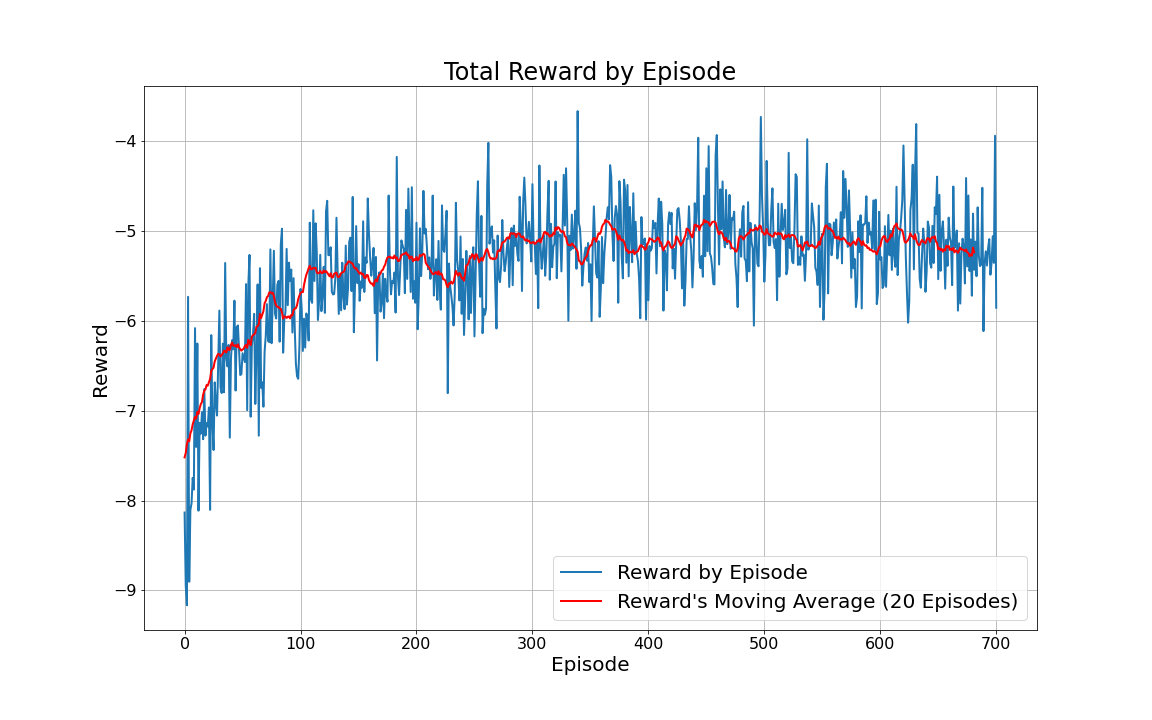}}
    \caption{Reward Curve of DDPG in each training episode.}
    \label{fig:Rewards}
\end{figure}


\section{Results and Analysis}\label{sec:Results}
\subsection{Qualitative Analysis}
In our experiment, we use the same model weights trained on Class-E Road profiles and test it on completely novel scenarios. The model is evaluated on three distinct test scenarios: (1) Simple-Bump Excitation; (2) Multiple-Hump Excitation; and (3) Stochastic ISO 8608 Class-E Road Profile (different from training set). Vehicle vertical body displacement, velocity, acceleration, and oscillation suppression time were analyzed to evaluate the car's movement and passengers' ride comfort.

The dynamic parameter of body displacement ($x_b$) indicates whether the vehicle excessively deviates from its stabilized position, resulting in unsafe maneuvers and uncomfortable riding experiences. The body velocity ($\dot{x}_b$) and acceleration ($\ddot{x}_b$) reflect the movement felt by the passengers. Higher body velocity and acceleration mean stronger and more abrupt impacts on the passengers, reducing their riding comfort. Therefore, these 3 quantities should be equally minimized to ensure both vehicle stability and passenger riding comfort.

After exiting the road roughness, the suspension system oscillates being fully stabilized, requiring a short period of time to suppress this oscillation. The time period measured from the end of the hump ($1500^{th}$ ms in Figure~\ref{fig:SingleHump}) until the vehicle body displacement is fully stabilized represents the time required for system stabilization. A shorter stabilization time indicates higher vehicle stability and better riding convenience.


\subsubsection{Performance on Single Bump}

\begin{figure*}[htbp]
  \centering
  \begin{tabular}{cc}
    \includegraphics[width=0.5\linewidth]{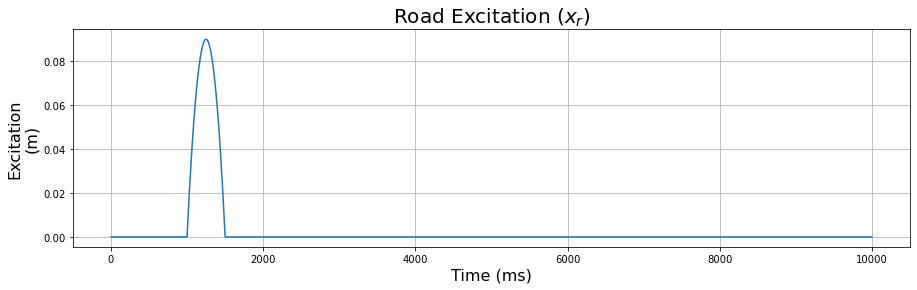} &
    \includegraphics[width=0.5\linewidth]{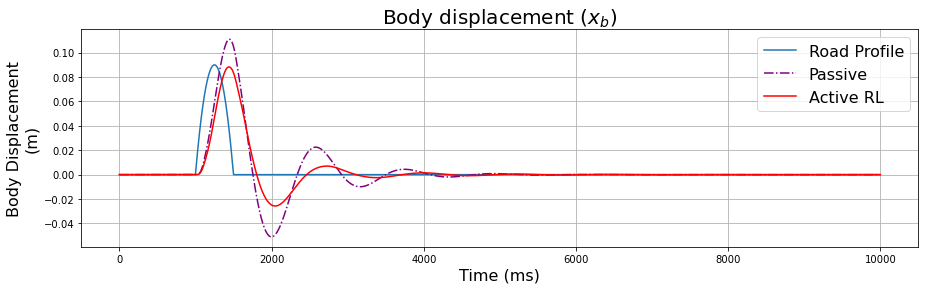} \\
    \includegraphics[width=0.5\linewidth]{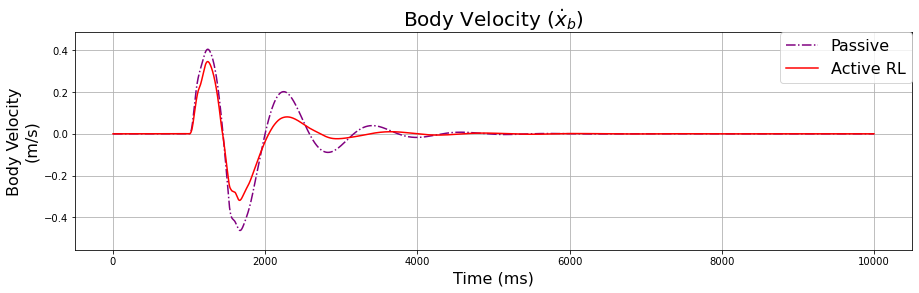} &
    \includegraphics[width=0.5\linewidth]{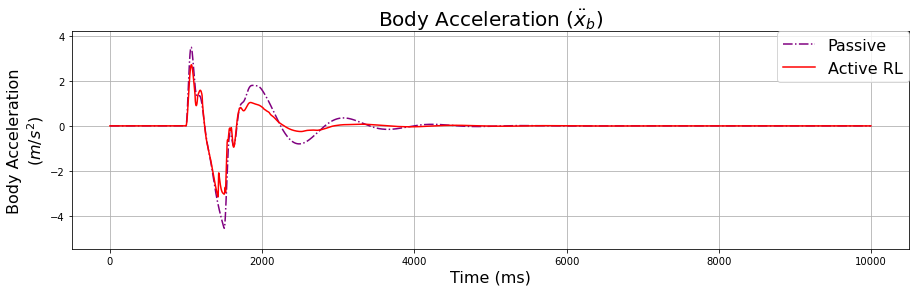} \\
    \includegraphics[width=0.5\linewidth]{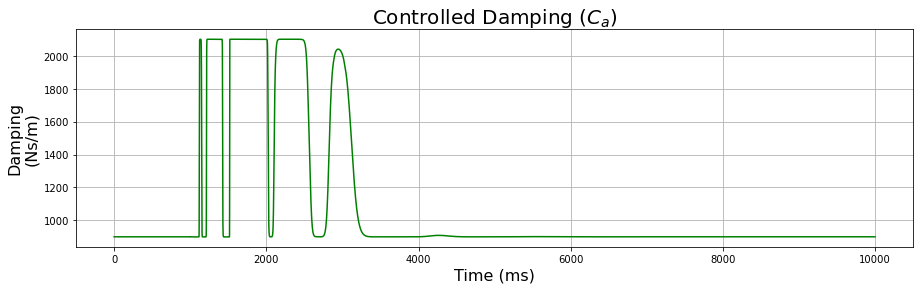} &
    \includegraphics[width=0.5\linewidth]{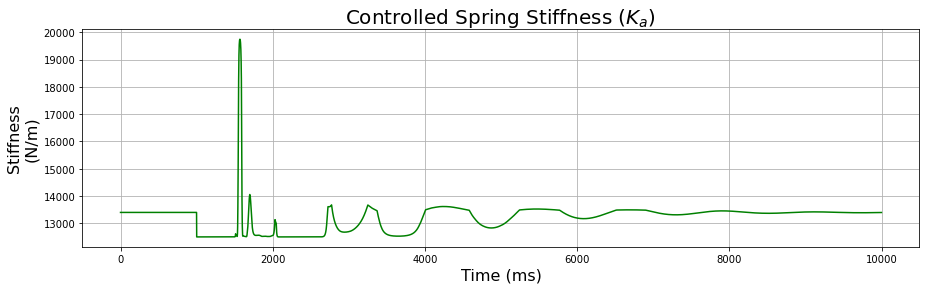} \\
  \end{tabular}
  \caption{Simple-Bump road: comparative performance of DRL-based Active Controller against Passive Suspension on a simple road bump, modeled with a sine curve. The road bump profile analysis was included for easy verification of the system's physical correctness. Graphs in the first row depict (1) A single hump road excitation and (2) Body Displacement with Passive and DRL Active suspension systems. A smoother displacement curve represents better driving stability. The second row displays the vehicle's Body Velocity (3) and Acceleration (4), respectively, under passive and active DRL system control. A velocity/acceleration closer to 0 represents smoother body movements and a better driving experience. The last two graphs illustrate the dynamic damping stiffness (5) and spring stiffness (6) controlled by DRL.}
  \label{fig:SingleHump}
\end{figure*}

\begin{figure*}[htbp]
  \centering
  \begin{tabular}{cc}
    \includegraphics[width=0.5\linewidth]{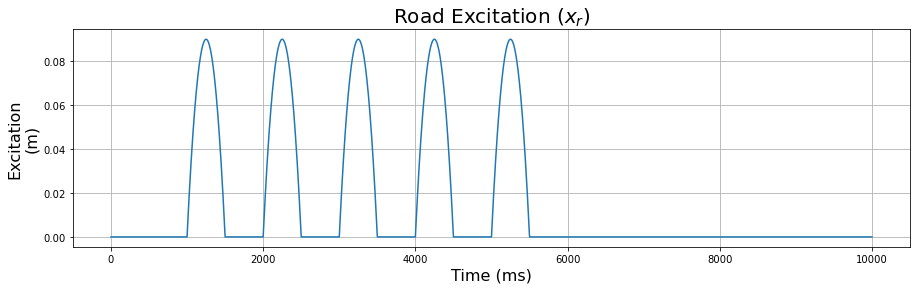} &
    \includegraphics[width=0.5\linewidth]{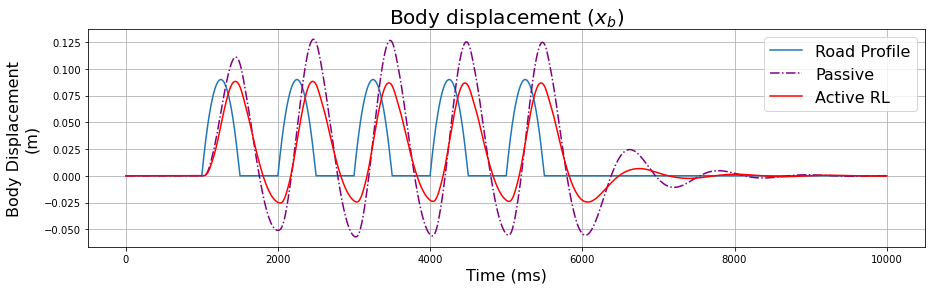} \\
    \includegraphics[width=0.5\linewidth]{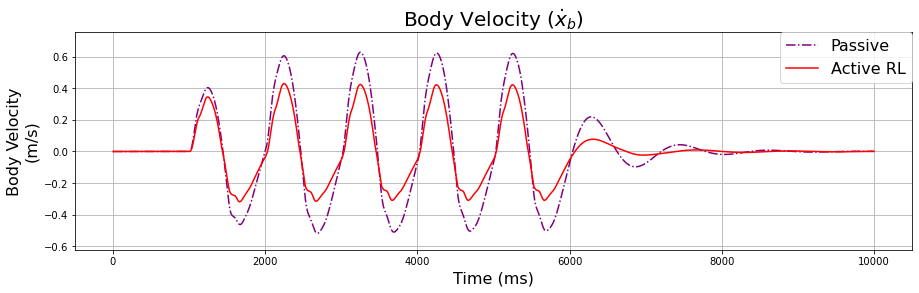} &
    \includegraphics[width=0.5\linewidth]{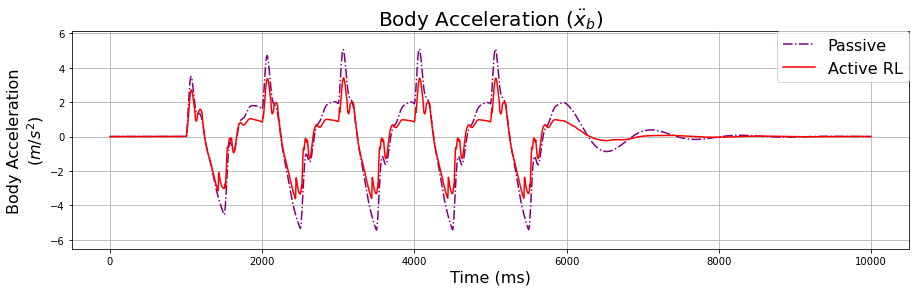} \\
    \includegraphics[width=0.5\linewidth]{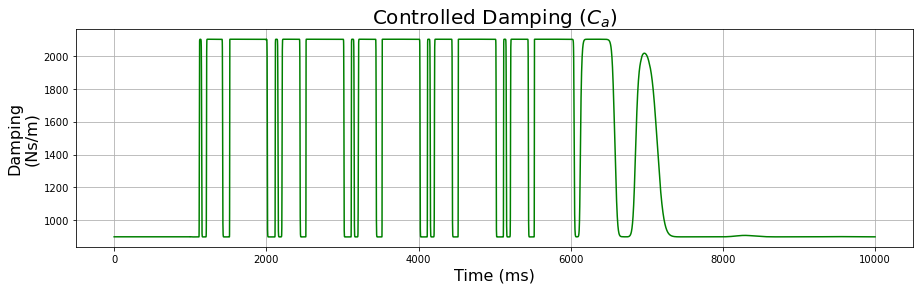} &
    \includegraphics[width=0.5\linewidth]{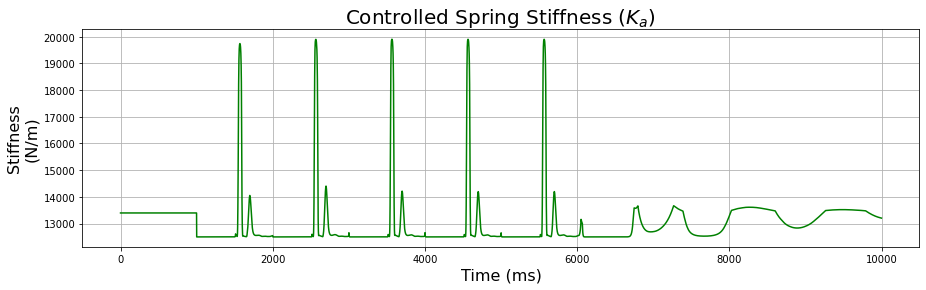} \\
  \end{tabular}
  \caption{Multiple Speed Humps road: comparative performance of DRL-based Active Controller against Passive Suspension. Similar to the simple single-bump profile, this multi-hump road profile is included for intuitive qualitative analysis.}
  \label{fig:MultiHump}
\end{figure*}
\begin{figure*}[htbp]
  \centering
  \begin{tabular}{cc}
    \includegraphics[width=0.5\linewidth]{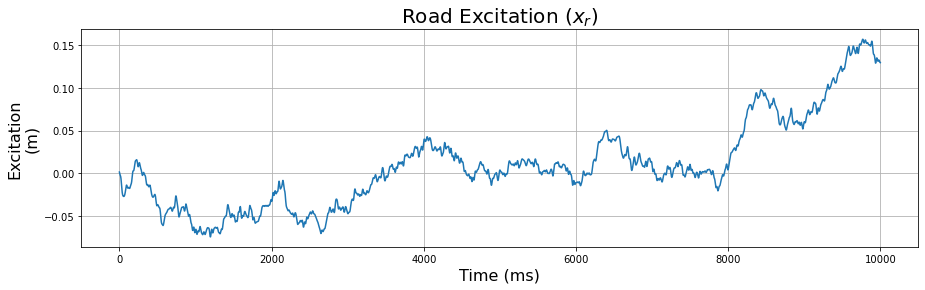} &
    \includegraphics[width=0.5\linewidth]{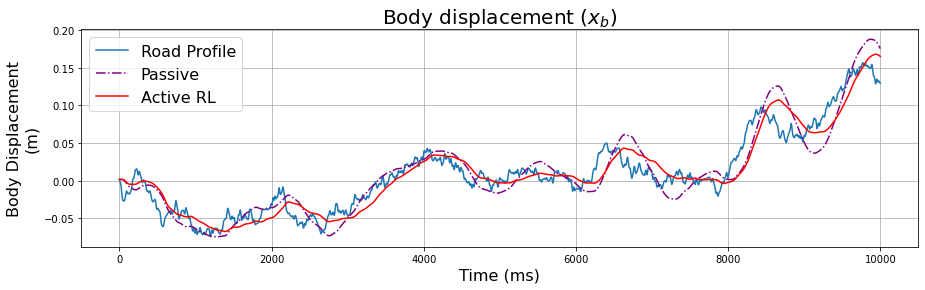} \\
    \includegraphics[width=0.5\linewidth]{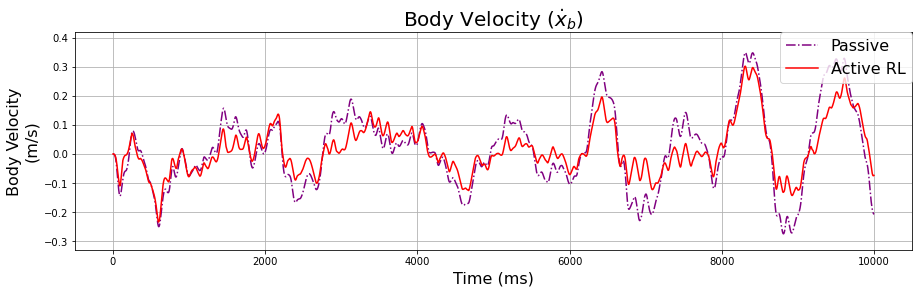} &
    \includegraphics[width=0.5\linewidth]{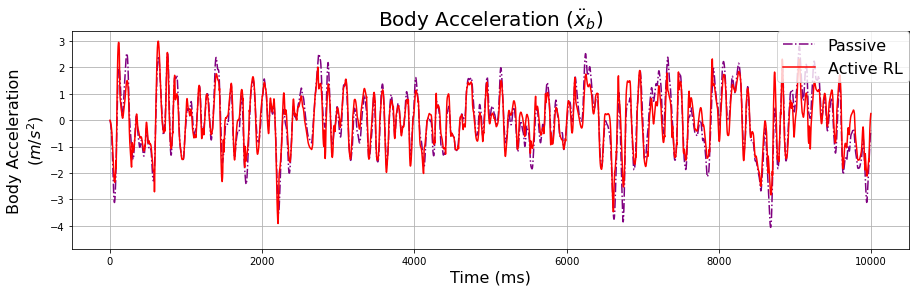} \\
    \includegraphics[width=0.5\linewidth]{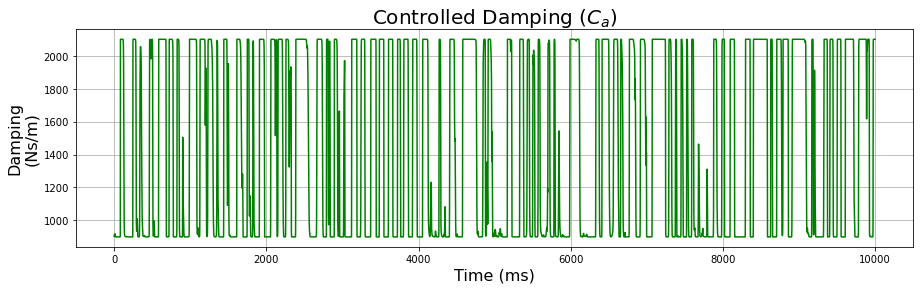} &
    \includegraphics[width=0.5\linewidth]{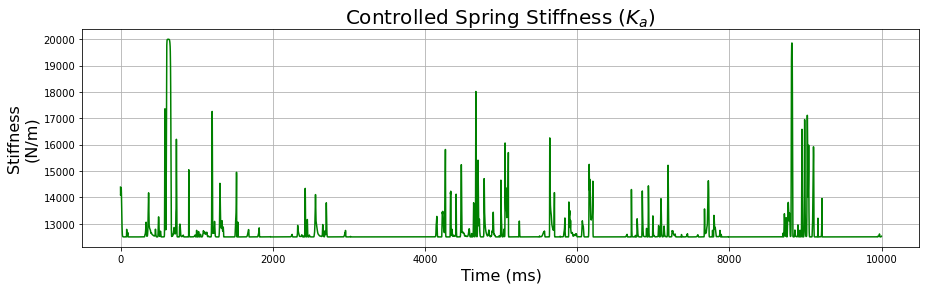} \\
  \end{tabular}
  \caption{ISO 8608 E-class stochastic road profile: comparative performance between DRL-based Active Suspension and Passive Suspension. The organization of the graphs is similar to that of the Simple-hump Road profile in Figure~\ref{fig:SingleHump}. Under the ISO road condition, a smoother, less oscillating body displacement curve means higher passenger riding comfort.}
  \label{fig:ISO-E}
\end{figure*}

Figure~\ref{fig:SingleHump} indicates suspension characteristics plotted as the vehicle crossed the single bump. Vehicle body displacement and acceleration were reduced compared to a passive suspension. The amplitude of these parameters from the active system was smaller while trade-off shorter settling time.



\subsubsection{Performance on Multiple Speed Humps}
When the vehicle crosses through the speed hump, the amplitude and frequency of the body vehicle displacement and acceleration, shown in Figure~\ref{fig:MultiHump} are much smaller than the passive system and stable during travel, which means the proposed suspension system demonstrates that the ride comfort and road-holding are improved, and eliminated the resonance of the wheel vibration to the body vibration. It shows that the system worked robustly with disturbances and uncertainties from the road.




\subsubsection{Performance on ISO 8608 Road Profile}



The comparison of vehicle dynamic performance on the ISO 8608 and E-class road profiles clearly shows that the active suspensions outperform the passive suspensions throughout the entire trip.  This indicates optimized ride comfort and road-holding in the active system, with body displacement and acceleration reduced by 41\% and 10.32\% compared to the passive system, respectively. Its dynamic response via Figure~\ref{fig:ISO-E} is smoother than the passive system, which provides a sense of comfort for passengers.

\subsection{Quantitative Analysis}
\begin{table}[htbp]
\caption{Comparison of DRL-controlled Active Suspension with Passive Suspension System}
\centering
\begin{tabular}{|c|c|c|c|}
    \hline
    \textbf{Evaluation metric}&\multicolumn{3}{|c|}{\textbf{Road Profile}} \\
    \cline{2-4} 
    & \textbf{\textit{Single-hump}}& \textbf{\textit{Multi-hump}}& \textbf{\textit{ISO E-class}} \\
    \hline
    \makecell{\% Velocity Reduction \\ (Overall)} & -44.4\% & -48.12\% & -43.58\% \\
    \hline
    \makecell{\% Velocity Reduction \\ (Q3 - high fluctuation)} & -60.76\% & -48.59\% & -45.06\% \\
    \hline
    \makecell{\% Acceleration Reduction \\ (Overall)} & -43.67\% & -57.08\% & -17.22\% \\
    \hline
    \makecell{\% Acceleration Reduction \\ (Q3 - high fluctuation)} & -69.82\% & -45.08\% & -18.62\% \\
    \hline
\end{tabular}
\label{Table:Quantitative}
\end{table}
The results presented in Table~\ref{Table:Quantitative} describes the quantitative measures obtained from three simulation scenarios. To assess passenger comfort improvements, 4 quantitative evaluation metrics have been selected, which include: the overall velocity reduction ($v_{overall}$) and Q3 velocity reduction ($v_{Q3}$), as well as the overall acceleration reduction ($a_{overall}$) and Q3 acceleration reduction ($a_{Q3}$). These metrics directly demonstrate the comparative performance of DRL-controlled and passive suspension systems in improving overall ride experience. The Q3 reduction evaluates how the DRL controller addresses more extreme vertical movements (75 percentile) of the vehicle body. 

The overall velocity deduction is defined as:

\begin{equation}
v_{overall} = \frac{\mu(v)_{DRL} - \mu(v)_{passive}}{\mu(v)_{passive}} \times 100\%
\end{equation}
Where $\mu(v)_{passive}$ and $\mu(v)_{DRL}$ are the mean absolute vertical body velocities for the passive and DRL-controlled systems, respectively:
\\
\begin{equation}
\mu(v)_{passive} = \frac{1}{K} \cdot \frac{1}{T} \sum_{k=1}^{K} \sum_{t=1}^{T} |v_{passive}|
\end{equation}

\begin{equation}
\mu(v)_{DRL} = \frac{1}{K} \cdot \frac{1}{T} \sum_{k=1}^{K} \sum_{t=1}^{T} |v_{DRL}|
\end{equation}
Where $K=50$ represents the number of experiments, and $T=10000$ represents the number of time steps per simulation. The variables $v_{DRL}$ and $v_{passive}$ represent the vertical body velocities at specific time stamps on specific experiments for the DRL-controlled and passive systems, respectively. $\mu(v)_{DRL}$ and $\mu(v)_{passive}$ are the mean absolute vertical body velocity of DRL-controlled or passive system. The overall acceleration reduction $a_{overall}$ is similarly defined.







For both acceleration and velocity reduction, Q3 reductions focus specifically on the top 25\% highest vertical velocity/acceleration to assess DRL performance in extreme scenarios where the vibration is particularly strong. 

The results demonstrate that the DRL achieves substantial reductions in body velocity and acceleration across three different road conditions. The DRL-controlled system shows a significant enhancement in both overall and Q3 velocity reduction, surpassing 40\%. Additionally, for single and multi-hump road profiles, the overall and Q3 acceleration reductions are also impressive, remaining above 40\%.
However, it's worth noting that on smoother roads with more frequent random small humps that are challenging to adjust, the DRL-controlled system achieves an approximate 18\% reduction in Q3/overall acceleration. This could be attributed to the smaller body acceleration resulting from the smoother road conditions compared to speed-humps.

In summary, the DRL-controlled system demonstrates remarkable performance in reducing body velocity and acceleration on different road profiles, achieving over 43\% reductions in both overall velocity and acceleration as well as reducing extreme vibration by $17\%$.


\section*{Conclusions}

In this paper, we present an innovative and physically realistic approach to active suspension control utilizing the DDPG algorithm. Specifically, the suspension system's damping and stiffness coefficients are considered 2 separate outputs of the Policy Network. These two variables are bounded within a realistic range and dynamically controlled to maximize the ride comfort. To facilitate the optimization of the control strategy, we implement a physical feedback environment based on a 2-DOF quarter suspension active suspension system model. As part of the environment, real-time road profiles are modeled based on ISO 8806 standards and are collected for training and validation. The obtained performance indicators of the vehicle serve as state inputs for reinforcement learning. This process involves continuously refining the control strategy based on reward and punishment functions and adjusting the weight matrix under varying road conditions until convergence is achieved.\\

Simulation results demonstrate that the proposed approach outperforms passive suspension systems in terms of vehicle ride comfort and road-holding. Vehicle body displacement and vertical acceleration reflect the dynamic response of the vehicle's suspension system to road disturbances. Qualitatively, compared to the Passive system, the DDPG-based controller has significantly lower vehicle body vibrations and oscillations. Quantitatively, the body velocity and acceleration are reduced by 20\% to 70\%. These statistics indicate minimized impacts felt by the passengers and increased vehicle stability, all of which are critical for safety and a better driving experience.

In future studies, we aim to further extend and investigate the applicability of our framework to half and full-vehicle suspension systems. These systems are more complex than the Quarter Car model due to the non-linear interactions between the wheels, introducing additional state variables and controllable coefficients. This poses a higher level of challenge for Deep Reinforcement Learning optimization and convergence, but it is a necessary milestone toward real-world deployment on commercial vehicles.

\printbibliography

@article{THEUNISSEN2021206,
title = {Preview-based techniques for vehicle suspension control: a state-of-the-art review},
journal = {Annual Reviews in Control},
volume = {51},
pages = {206-235},
year = {2021},
issn = {1367-5788},
doi = {https://doi.org/10.1016/j.arcontrol.2021.03.010},
url = {https://www.sciencedirect.com/science/article/pii/S1367578821000183},
author = {Johan Theunissen and Antonio Tota and Patrick Gruber and Miguel Dhaens and Aldo Sorniotti}
}

@article{review_suspension,
author = {Jiregna, Iyasu and Sirata, Goftila Gudeta},
year = {2020},
month = {11},
pages = {109-114},
title = {A review of the vehicle suspension system Iyasu JIREGNA, Gof tila SIRATA A REVIEW OF THE VEHICLE SUSPENSION SYSTEM},
volume = {4},
journal = {Journal of Mechanical and Energy Engineering},
doi = {10.30464/jmee.2020.4.2.109}
}

@article{doi:10.1177/1461348419876392,
author = {AMA Soliman and MMS Kaldas},
title ={Semi-active suspension systems from research to mass-market – A review},
journal = {Journal of Low Frequency Noise, Vibration and Active Control},
volume = {40},
number = {2},
pages = {1005-1023},
year = {2021},
doi = {10.1177/1461348419876392},
URL = {https://doi.org/10.1177/1461348419876392},
eprint = {https://doi.org/10.1177/1461348419876392}
}

@article {10.6703--570436548,
title = {Study on the sliding mode control method for the active suspension system},
author = {Nguyen, Tuan Anh},
volume = 18,
issue = 5,
year = 2021,
pub_date = {2021-09-01},
pages = {1--10},
doi = {10.6703/IJASE.202109_18(5).006},
url = {https://doi.org/10.6703/IJASE.202109_18(5).006},
journal = {International Journal of Applied Science and Engineering},
issn = {1727-7841},
publisher = {Chaoyang University of Technology}
}

@INPROCEEDINGS{10072041,
  author={Xia, Xiangjun and Ning, Donghong and Liao, Yulin and Liu, Pengfei and Du, Haiping and Li, Weihua},
  booktitle={2022 IEEE PES 14th Asia-Pacific Power and Energy Engineering Conference (APPEEC)}, 
  title={A Novel Semi-Active Electromagnetic Suspension for Ride Comfort and Energy Harvesting}, 
  year={2022},
  volume={},
  number={},
  pages={1-7},
  doi={10.1109/APPEEC53445.2022.10072041}}

@article{10.1155/2023/2889435,
author = {Ngoc, Nguyen and Nguyen, Tuan Anh},
year = {2023},
month = {02},
pages = {},
title = {The Dynamic Model and Control Algorithm for the Active Suspension System},
volume = {2023},
journal = {Mathematical Problems in Engineering},
doi = {10.1155/2023/2889435}
}

@article{MOHITE20184317,
title = {Development of Linear and Non-linear Vehicle Suspension Model},
journal = {Materials Today: Proceedings},
volume = {5},
number = {2, Part 1},
pages = {4317-4326},
year = {2018},
note = {7th International Conference of Materials Processing and Characterization, March 17-19, 2017},
issn = {2214-7853},
doi = {https://doi.org/10.1016/j.matpr.2017.11.697},
url = {https://www.sciencedirect.com/science/article/pii/S221478531732970X},
author = {Ajit G. Mohite and Anirban C. Mitra}
}

@INPROCEEDINGS{5459681,
  author={Tang, Chuan Yin and Zhao, Guang Yao and Zhou, Wei and Zhou, Shu Wen},
  booktitle={2010 International Conference on Measuring Technology and Mechatronics Automation}, 
  title={Research on Suspension System Based on Neural Network Algorithm}, 
  year={2010},
  volume={3},
  number={},
  pages={186-189},
  doi={10.1109/ICMTMA.2010.188}}

@INPROCEEDINGS{8606740,
  author={Chen, Hsin-Chang and Lin, Yu-Chen and Chang, Yu-Heng},
  booktitle={2018 International Automatic Control Conference (CACS)}, 
  title={An Actor-Critic Reinforcement Learning Control Approach for Discrete-Time Linear System with Uncertainty}, 
  year={2018},
  volume={},
  number={},
  pages={1-5},
  doi={10.1109/CACS.2018.8606740}}

@INPROCEEDINGS{9442091,
  author={Wang, Xi and Zhuang, Weichao and Yin, Guodong},
  booktitle={2020 IEEE 18th International Conference on Industrial Informatics (INDIN)}, 
  title={Learning-Based Vibration Control of Vehicle Active Suspension}, 
  year={2020},
  volume={1},
  number={},
  pages={94-99},
  doi={10.1109/INDIN45582.2020.9442091}}

@ARTICLE{8950280,
  author={Ming, Liu and Yibin, Li and Xuewen, Rong and Shuaishuai, Zhang and Yanfang, Yin},
  journal={IEEE Access}, 
  title={Semi-Active Suspension Control Based on Deep Reinforcement Learning}, 
  year={2020},
  volume={8},
  number={},
  pages={9978-9986},
  doi={10.1109/ACCESS.2020.2964116}}

@book{handbook,
    author = {Jiangtao Caoa,c, Ping Lib and Honghai Liuc.},
    title = {Handbook of Vehicle Suspension Control Systems},
    publisher = {The Institution of Engineering and Technology},
    year = {2014},
    chapter = {2}
}

@techreport{ISO8608,
  author = {ISO (International Organization for Standardization)},
  title = {Mechanical Vibration - Road Surface Profiles - Reporting of Measured Data},
  institution = {{TC108/SC2 (Mechanical Vibration and Shock/Measurement and Evaluation of Mechanical Vibration and Shocks as Applied to Machines, Vehicles and Structures)}},
  year = {2016},
  type = {ISO Standard},
  number = {ISO 8608: E},
}

@misc{1509.02971,
Author = {Timothy P. Lillicrap and Jonathan J. Hunt and Alexander Pritzel and Nicolas Heess and Tom Erez and Yuval Tassa and David Silver and Daan Wierstra},
Title = {Continuous control with deep reinforcement learning},
Year = {2015},
Eprint = {arXiv:1509.02971},
doi = {10.48550/arXiv.1509.02971}
}

@article{KOBAYASHI202163,
title = {t-soft update of target network for deep reinforcement learning},
journal = {Neural Networks},
volume = {136},
pages = {63-71},
year = {2021},
issn = {0893-6080},
doi = {https://doi.org/10.1016/j.neunet.2020.12.023},
url = {https://www.sciencedirect.com/science/article/pii/S0893608020304482},
author = {Taisuke Kobayashi and Wendyam Eric Lionel Ilboudo}
}

@article{DBLP:journals/corr/MnihKSGAWR13,
  author       = {Volodymyr Mnih and
                  Koray Kavukcuoglu and
                  David Silver and
                  Alex Graves and
                  Ioannis Antonoglou and
                  Daan Wierstra and
                  Martin A. Riedmiller},
  title        = {Playing Atari with Deep Reinforcement Learning},
  journal      = {CoRR},
  volume       = {abs/1312.5602},
  year         = {2013},
  url          = {http://arxiv.org/abs/1312.5602},
  eprinttype    = {arXiv},
  eprint       = {1312.5602},
  bibsource    = {dblp computer science bibliography, https://dblp.org}
}

@phdthesis{shen2022,
  author = {Shen, Daoyu},
  title = {A Study on Active Suspension System with Reinforcement Learning},
  school = {University of Technology Sydney},
  address = {Australia},
  year = {2022},
  type = {Ph.D. thesis},
  number = {30610651}
}

@misc{babaahmadi2023deep,
      title={A Deep Reinforcement Learning-Based Controller for Magnetorheological-Damped Vehicle Suspension}, 
      author={AmirReza BabaAhmadi and Masoud ShariatPanahi and Moosa Ayati},
      year={2023},
      eprint={2301.02714},
      archivePrefix={arXiv},
      primaryClass={eess.SY}
}
\end{document}